\documentclass[letterpaper]{article} 
\usepackage{aaai2026}  
\usepackage{times}  
\usepackage{helvet}  
\usepackage{courier}  
\usepackage[hyphens]{url}  
\usepackage{graphicx} 
\usepackage{natbib}  
\usepackage{caption} 
\usepackage{algorithm}
\usepackage{algorithmic}

\usepackage{booktabs}

\usepackage{newfloat}
\usepackage{listings}
\DeclareCaptionStyle{ruled}{labelfont=normalfont,labelsep=colon,strut=off} 
\floatstyle{ruled}
\newfloat{listing}{tb}{lst}{}
\floatname{listing}{Listing}
\title{LASTIST: LArge-Scale Target-Independent STance dataset}
\author{
    DongJae Kim\textsuperscript{\rm 1},
    Yaejin Lee\textsuperscript{\rm 1},
    Minsu Park\textsuperscript{\rm 1},
    Eunil Park\textsuperscript{\rm 1}
}
\affiliations{
    \textsuperscript{\rm 1}Sungkyunkwan University\\
    bronze.ash@g.skku.edu, 
    19lyaejin@gmail.com,
    alstn7@gmail.com,
    eunilpark@skku.edu
}

\usepackage{bibentry}

\usepackage{times}
\usepackage{helvet}
\usepackage{courier}
\usepackage{xcolor}
\begin{document}

\maketitle

\begin{abstract}
Stance detection has emerged as an area of research in the field of artificial intelligence. However, most research is currently centered on the target-dependent stance detection task, which is based on a person's stance in favor of or against a specific target. Furthermore, most benchmark datasets are based on English, making it difficult to develop models in low-resource languages such as Korean, especially for an emerging field such as stance detection. This study proposes the LArge-Scale Target-Independent STance (LASTIST) dataset to fill this research gap. Collected from the press releases of both parties on Korean political parties, the LASTIST dataset uses 563,299 labeled Korean sentences. We provide a detailed description of how we collected and constructed the dataset and trained state-of-the-art deep learning and stance detection models. Our LASTIST dataset is designed for various tasks in stance detection, including target-independent stance detection and diachronic evolution stance detection. We deploy our dataset on \url{https://anonymous.4open.science/r/LASTIST-3721/}.
\end{abstract}


\section{Introduction}

Stance Detection (SD) is a newly emerging area in the artificial intelligence research domain.
As seen in the SemEval-2016 dataset, stance detection has been mainly formulated as a classification problem for given sentences into Favor, Against optionally neutral or none toward a specific target~\cite{mohammad2016semeval}.
Backed by the growing interest in web technology and social media platforms, the number of studies focusing on the stance detection problem has increased in recent years~\cite{kuccuk2020stance, alturayeif2023systematic, gera2025deep}.

Although stance detection problems have become one of the most prominent problems in Machine Learning (ML), stance detection studies today share several commonalities that hinder their performance.
First, stance detection datasets constructed for model development tend to possess a limited number of instances, making it difficult to train a model that can accurately classify stance.
A recent survey paper on stance detection~\cite{kuccuk2020stance, alturayeif2023systematic} revealed that the majority of studies consist of a limited number of data instances, varying from 1k to 50k, with the exception of~\citet{10.1007/978-3-030-30179-8_4}, which has 387k.
Compared to datasets in other Natural Language Processing (NLP) fields, such as sentiment analysis~\cite{JIM2024100059} or question answering~\cite{10.1145/3560260}, the insufficiency of large-scale datasets contributes to the lack of performance of stance detection models.

Furthermore, stance detection models tend to be target-specific because of their formulation methodology, which means that they are bound to one or more specific targets.
While some datasets incorporate multiple targets to mitigate this limitation, well-known datasets on stance detection tasks are usually bound to the particular events such as the U.S. presidential election~\cite{10.1007/978-3-031-72241-7_1, sobhani-etal-2017-dataset, li2021p, niu2024challenge}, major societal issues such as the feminist movement or climate change~\cite{mohammad2016semeval, 10.1145/3003433, 10.1007/978-3-031-78538-2_28, upadhyaya2023multi, wang2024multiclimate}.
Although this tendency could be effective when predicting the stance of a tweet or an article for a specific topic, it leads to another problem of target dependency.
Because the models are bound to specific targets, their performance toward unseen targets or a universal stance could be limited.
This problem could be even more critical in the case of NLP for low-resource languages such as Korean, as the resources for training ML models are even scarcer.

To overcome this research gap, we propose the LArge-Scale Target-Independent STance (LASTIST) dataset, which is a large-scale dataset that can be used in the Korean stance detection task.
This study makes the following contributions:
1) We present LATIST, a large-scale Korean stance detection dataset consisting of 563,299 sentences that can be classified as pro-left or pro-right.
2) We propose a framework for dataset construction based on active learning, which suggests an efficient means of extracting biased sentences from press releases.
3) We conduct several experiments on our LASTIST dataset, including benchmark and Large Language Models (LLMs), which provide a suitable benchmark for further stance detection research.

The remainder of this paper is organized as follows. In Section 2, we present previous studies that attempted to build stance detection datasets, classifying them into single-target-specific, multi-target-specific, and target-independent datasets. Section 3 provides a detailed description of how we collected the data and filtered the properly biased expressions.
Section 4 reports the experimental results using baselines and LLMs on the stance detection dataset.
Finally, we conclude the paper by stating the implications and limitations of our work.

\section{Related Works}

Stance detection tasks typically classify the viewpoint expressed in a text toward one or more specific targets, which refers to an entity or short noun-phrase that a given document or sentence aims to express opinion or attitude, for example, a political figure or controversial topic ~\cite{zotova-2020, conforti-2020}. 
Depending on the nature or number of targets involved, stance detection studies can be classified into three categories: single-target, multi-target, and target-independent.

\subsection{Single-Target Stance Detection}
Single-target stance detection refers to predicting the stance expressed in a given text toward a predefined single target. 
Owing to the ease of the construction approach, prior studies mostly centered on single-target stance detection, and most existing datasets have been designed accordingly~\cite{alturayeif2023systematic, ALDAYEL2021102597}.
One of the most famous examples is \citet{mohammad2016semeval}, which introduced an English dataset on topics such as atheism or climate change, aiming to identify rumors and the stance of Twitter users through their textual replies. 
More recently, \citet{li2021p} presented a single-target stance detection dataset for several U.S. presidential candidates in the political domain, composed of 21,574 tweets.

Although such datasets are effective for training models that can classify stances toward a specific target, several challenges often arise while building them.
First, retaining sufficient annotated data for each specific target is challenging, which leads to datasets lacking the size needed to effectively train deep learning models, such as LLMs.
As shown in Table~\ref{tab:stance_datasets}, most target-specific datasets consist of up to 50k data instances, making it difficult for the stance detection model to be trained without additional training methods, such as fine-tuning.
Moreover, the model's narrow focus on single-target approaches hinders its ability to learn the underlying relationships among targets, capture domain-general features, and limit generalization to unseen targets. 
Finally, a notable limitation of existing resources is their dominant focus on English-language data, which highlights a significant resource gap for other languages, such as Korean~\cite{mohammad2016semeval, conforti-2020, grimminger-2021, gyawali-2024, ezstance-2024, li-2024}.

\subsection{Multi-Target Stance Detection}
These drawbacks of single-target approaches have spurred interest in multi-target stance detection (MTSD) and the development of a corresponding dataset to support such a task. 
MTSD focuses on identifying stances toward multiple targets within the input text, thereby enabling a more comprehensive and realistic understanding of the expressed stance. 
One early effort in this direction was the Multi-Target SD dataset~\cite{sobhani-etal-2017-dataset}, which simultaneously introduced annotations for two U.S. political targets.
Similarly, the Trump-Hillary dataset~\cite{darwish-2017} provided multi-target annotations for both U.S. presidential candidates--Hillary and Trump as targets--for instance, “supporting Hillary” and “opposing Trump”.  
More recently, \citet{niu2024challenge} constructed a stance detection dataset based on conversational data to address real-world applications. 
Such datasets effectively ascertain the practical challenges of MTSD, including coreference relations or implicit target references that frequently occur in the real world.
Despite its relevance, MTSD has received limited attention in recent years owing to the scarcity of high-quality datasets containing multiple annotated targets. 
Moreover, most MTSD datasets are constructed in English, further highlighting the lack of resources available for other languages. 

\subsection{Target-Independent Stance Detection}
Target-independent stance detection aims to identify the stance expressed in a text without relying on explicit or predefined target entities, thereby hindering the constraint on the number of target types.
The IBM Debater~\cite{bar-2017} dataset leveraged Wikipedia articles and was labelled with stances on 55 claims, making it compatible with target-independent tasks. 
Similarly, RumourEval-19~\cite{gorrell-2019} extended stance detection to cover different topic entities as target-independent entities related to natural disasters from Tweets and Reddit posts.  

\begin{table*}[t]
\centering
\footnotesize
\begin{tabular}{l c r c r c}
\toprule
\textbf{Dataset} & \textbf{Target} & \textbf{\# Target(s)} & \textbf{Source} & \textbf{Dataset Size} & \textbf{Language} \\
\midrule
SemEval-2016 Task 6 ~\cite{mohammad2016semeval} & ST & 6 & Twitter & 4,870 & English \\
P-stance ~\cite{li2021p} & ST & 3 & Twitter & 21,574 & English \\
WT-WT ~\cite{conforti-2020} & ST & 5 & Twitter & 51,284 & English \\
EZ-Stance ~\cite{ezstance-2024} & ST & 40,678 & Twitter & 47,316 & English \\

Multi-Target SD ~\cite{sobhani-etal-2017-dataset} & MT & 4 & Twitter & 4,455 & English \\
Trump-Hillary ~\cite{darwish-2017} & MT & 2 & Twitter & 3,450 & English \\
MT-CSD ~\cite{niu2024challenge} & MT & 5 & Reddit & 15,876 & English \\
C-Stance ~\cite{cstance-2023} & MT & 7 & Weibo & 48,126 & Chinese \\

IBM Debater ~\cite{bar-2017} & TI & 55 & Wikipedia & 2,934 & English \\
RumourEval-19 ~\cite{gorrell-2019} & TI & 8,574 & Twitter, Reddit & 8,574 & English \\
ORCHID ~\cite{orchid-2023} & TI & 2,436 & Debate videos & 14,091 & Chinese \\
Arabic News Stance ~\cite{khouja-2020} & TI & 3,786 & News & 3,786 & Arabic \\
\midrule
\textbf{LASTIST (Ours)} & ST, TI & - & Press release articles & 563,299 & Korean \\
\bottomrule
\end{tabular}
\caption{Comparison of existing stance detection datasets across various targets and languages. \textit{Target} denotes the dependency of target (\textit{ST}: single-target, \textit{MT} : multi-target, \textit{TI} : target-independent)}
\label{tab:stance_datasets}
\end{table*}

To the best of our knowledge, there are currently no available Korean stance detection datasets that support target-independent tasks. 
Furthermore, the effectiveness of stance detection models heavily depends on access to sufficiently large and well-annotated datasets that are also balanced. 
Therefore, this study proposes the LASTIST dataset, aimed at supporting target-independent stance detection while ensuring a sufficient data size and extending accessibility to the low-resource language, Korean.

\section{Dataset Construction}
\begin{figure}[ht!]
    \centering
    \includegraphics[width=0.8\linewidth]{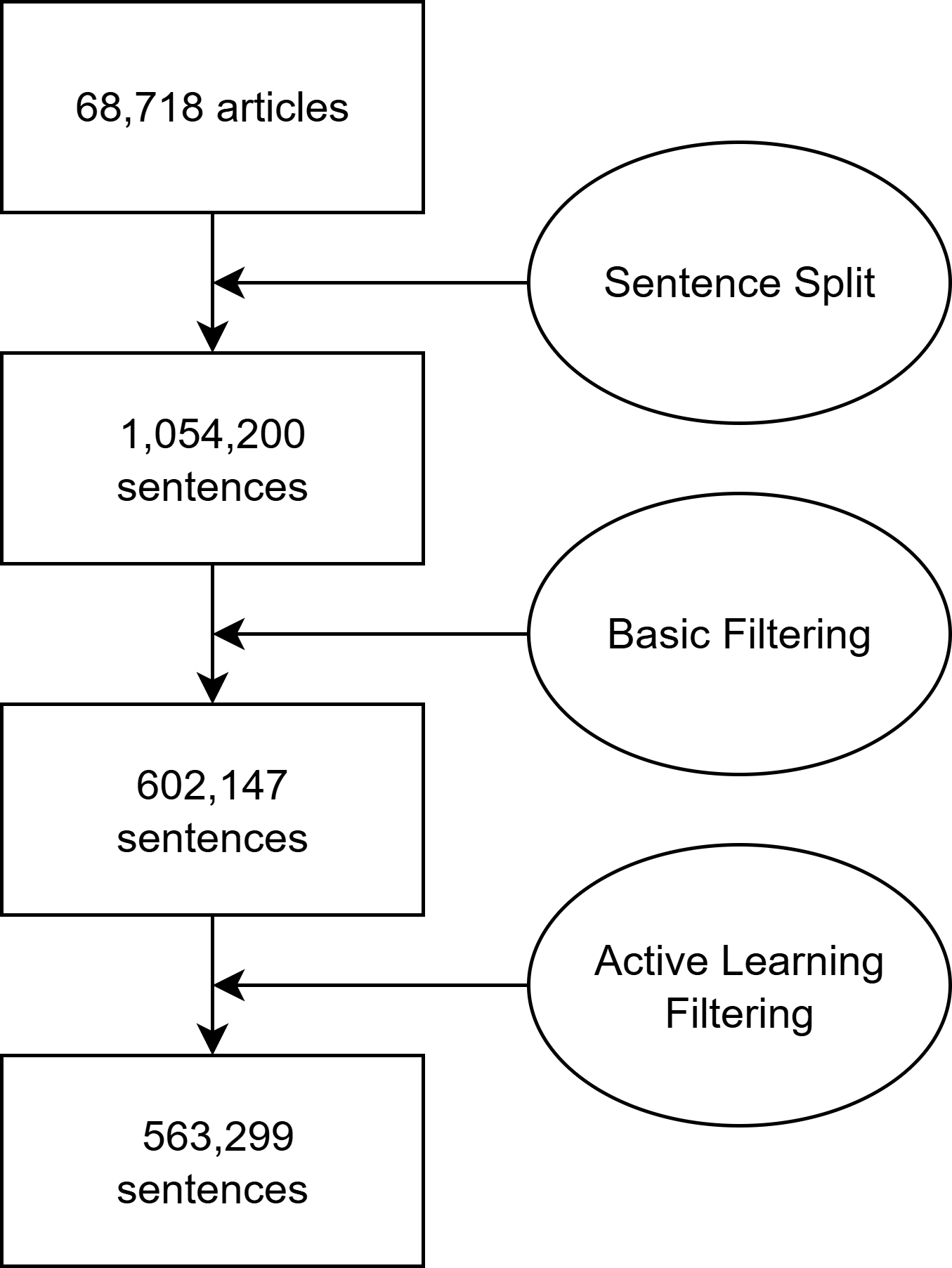}
    \caption{Dataset construction process.}
    \label{fig:figure1}
\end{figure}

This section describes how we collected the data to construct the LASTIST dataset and preprocessed it to retain only the sufficiently biased sentences.
This is followed by topic labeling and quality assurance of our dataset, which could be used in various stance detection tasks to work on further stance detection tasks.
Figure~\ref{fig:figure1} illustrates the overall dataset construction.

\subsection{Data Collection}
To collect the biased articles that could be classified into pro-left and pro-right, we collected the press release articles from the webpages of The Minjoo party~\footnote{\url{https://theminjoo.kr/main/}} and People Power party~\footnote{\url{https://www.peoplepowerparty.kr/}}.
Both parties are regarded as the pro-left and pro-right sides of the Korean political spectrum, and they are sufficient to be used as markers for Korean stance detection studies, as in previous studies on English-based stance detection tasks that tend to classify the labels as pro-left and pro-right~\cite{kim2025know, he-etal-2024-reading, 10.1007/978-3-031-70819-0_1}.
We used Selenium~\footnote{\url{https://www.selenium.dev/}} to crawl the content of press releases from the websites, and regarded the articles collected from the Minjoo party as pro-left, and those collected from the People Power party as pro-right.
From this procedure, we collected 21,378 pro-right and 47,340 pro-left articles for analysis.

\subsection{Data Preprocessing}
After collecting a dataset that would serve as a basis for our dataset, we preprocessed the data for our targeted sentence-level stance-detection task.
First, we split the documents collected into sentences.
This is because stances are typically expressed in specific sentences rather than across an entire document, making document-level labels prone to noise.
In addition, sentence-level inputs are better suited for pretrained models, such as BERT, and allow for more straightforward annotation guidelines, leading to higher consistency and better interpretability.
This preprocessing resulted in approximately 1,054,317 sentences, of which 774,327 were labeled pro-left and 279,990 pro-right.

We then filtered out data that were inappropriate for stance detection tasks.
We first excluded boilerplates, such as signature lines and publication dates.
Sentences that were excessively short or long for stance analysis were excluded, retaining only those between five and 30 words.
We obtained 602,147 sentences in total, of which 431,822 were labeled pro-left and 188,325 pro-right.

\subsection{Active Learning Filtering}
In this study, we assumed that subjective expressions inherently entail stances.
This is supported by the assumption that only sentences containing subjective content, such as opinions, evaluations, or emotions, can meaningfully convey a stance toward an arbitrary target.
As theorized in studies in linguistics, sociolinguistics, and humanities regarding stance detection~\cite {kockelman2004stance, du2008stance, mets2024automated}, subjectivity is not an independent concept from stance, but rather an indexical effect that emerges as a result of stance expression.
Therefore, subjectivity can be used as a valid indicator of the existence of a stance, which justifies the use of subjectivity to filter out non-stance-related sentences.

Therefore, we employed a subjectivity-based auto-labeling framework to refine the dataset by filtering out instances that were still misaligned with stance detection.
In particular, we used Small-Text~\cite{schroder-etal-2023-small}, an active learning framework designed to easily supplement labels for text classification tasks.
Based on Small-Text, we used KPF-BERT~\footnote{\url{https://github.com/KPFBERT/kpfbert}}, which is pre-trained by the Korean Press Foundation using BigKinds news data made up of Korean news articles. 
For the annotation guidelines, we used an excerpt from \citet{antici-etal-2024-corpus}, which conveys the definition of subjectivity in news articles.
We trained the KPF-BERT model using 10 iterations and filtered out sentences inappropriate for stance detection, such as quotations or factual sentences.
We applied this method because quotations are likely to convey the stance of the person referred to in the sentence, which might not align with the overall document's stance.

\subsection{Dataset Distribution}
Following the procedures mentioned above, we finally obtained 563,299 data points.
The final LASTIST dataset consists of 394,763 pro-left and 168,536 pro-right sentences.
The detailed statistics of the LASTIST dataset are presented in Table~\ref{tb:statistics}.

\begin{table}
    \centering
    \begin{tabular}{l c c }
        \toprule
        & Pro-Left & Pro-Right \\ \midrule
        Number of sentences & 394,763 & 168,536 \\
        Avg. \#Tokens & 15.734 & 15.274 \\
        Avg. Length & 68.054 & 63.159 \\\bottomrule
    \end{tabular}
    \caption{Dataset statistics for LASTIST.}
    \label{tb:statistics}
\end{table}

\section{Experiment}
\label{sec:experiment}
Following the construction of the LASTIST dataset, we conducted an empirical study to evaluate its effectiveness and validity by benchmarking a bidirectional encoder representation from transformers (BERT)-based stance detection method~\cite{devlin2019bert}. 
Our experiments aimed to validate LASTIST as a reliable benchmark suitable for conventional stance detection tasks and the more challenging target-independent context. 

\subsection{Experiment Setting}
To assess the validity of the LASTIST dataset across both target-independent and target-dependent stance detection tasks, we conducted experiments under two different settings: 

1) Target-Independent Setting (Full LASTIST): The complete LASTIST dataset encompasses a broad and diverse range of targets, positioning it as an appropriate benchmark for evaluating target-independent stance detection. In this setting, the entire dataset was leveraged without providing the model with explicit information regarding the target entities. This approach enabled a comprehensive evaluation of the dataset’s suitability for the target-independent stance detection task. 

2) Single-Target Setting (LASTIST subset): To further examine the dataset’s applicability in narrowly scoped scenarios, we conducted additional experiments focused on a single-target stance detection task, specifically centered on one target, party leadership. A refined subset of the LASTIST dataset was rendered using the Latent Dirichlet Allocation (LDA) topic modeling method to exclude irrelevant and dissimilar topics. Thus, this setting yielded a coherent dataset that evaluated the stance detection performance under constrained, single-target conditions.

\begin{table*}[ht]
\centering
\begin{tabular}{lccc}
\toprule
\textbf{Dataset} & \textbf{Pro-Left} & \textbf{Pro-Right} & \textbf{Total} \\
\midrule
Target-Independent & 394,273 & 168,536 & 563,299 \\
Single-Target & 67,584 (17.1\%) & 37,866 (22.5\%) & 105,450 (18.7\%) \\
\bottomrule
\end{tabular}
\caption{Size of dataset in target-independent and single-target settings}
\label{tab:subset}
\end{table*}

\subsection{Model Construction} 
In the context of the BERT-based baseline, our model comprises a pretrained KoBERT-based encoder to derive the initial token-level embeddings for each input sentence. 
These token embeddings are then aggregated into a fixed-size sentence embedding using a mean pooling layer. 

For the learning strategy, the model adopts a contrastive learning approach on these sentence embeddings to achieve the following two goals: 

\begin{enumerate}
    \item Encourage the model to capture stance-specified contextual features, and
    \item Effectively differentiates between sentences belonging to divergent stances by pulling sentence embeddings from the same label closer while pushing different labels away. 
\end{enumerate}

This contrastive learning framework fosters the sentence embeddings to become stance-discriminative, improving their utility for downstream stance detection tasks, such as political bias classification. 
Unlike a simple token combination, the pooling layer effectively aggregates token-level embeddings into a coherent sentence-level embedding. 
The resulting sentence embeddings are then passed into a simple one-layer classifier, which performs the final prediction of the political bias class.

Configuring the model, we jointly trained the contrastive learning loss and the political bias classification loss.
A weighting parameter $\alpha$ was introduced to balance the two objectives, which was impirically set to 0.5 to ensure the equal emphasis during training. 
The model was optimized using Adam with a batch size of 8 and 1e-5 learning rate. 
To prevent overfitting, a dropout rate of 0.5 was also applied to the classifier layer. 
For the Single-Target model, most hyperparameters remained consistent with the previous setup. 
However, to account for the smaller dataset size, the batch size was reduced to 4, the learning rate was adjusted to 2e-5, and the number of training epochs was set to 30.
All the experiments were performed on a single GPU, NVIDIA GeForce RTX 2080 Ti (11GB VRAM) under CUDA 11.6 and driver version 510.54, ensuring sufficient computational resources for BERT-based stance detection training.
Also, we fixed the random seed to 42 across all training and evaluation runs.

\subsection{Evaluation Metrics}
For our evaluation, we used three evaluation metrics to report the performance of the model. The evaluation metrics used were as follows:

\subsubsection{Accuracy}
We used accuracy as a metric to assess the overall performance and evaluate how well the model correctly classifies pro-left and pro-right sentences.
The calculation is as follows, where $T_{Left}$ and $T_{Right}$ denote the correct predictions for pro-left and pro-right sentences, respectively.
In addition, $F_{Left}$ and $F_{Right}$ denote incorrect predictions for pro-left and pro-right sentences, respectively.

\begin{equation}
Accuracy=\frac{T_{Left} + T_{Right}}{T_{Left} + T_{Right} + F_{Left} + F_{Right}}
\end{equation}

\subsubsection{F1-score}
We also used the macro-average of the F1-score to mitigate the effect of label imbalance. We first calculated the F1-score for each label and then computed the average as follows, where $P$ and $R$ refer to Precision and Recall, respectively.

\begin{equation}
F1_{Left}=\frac{2 \cdot P_{Left} \cdot R_{Left}}{P_{Left} + R_{Left}}
\end{equation}
\begin{equation}
F1_{Right}=\frac{2 \cdot P_{Right} \cdot R_{Right}}{P_{Right} + R_{Right}}
\end{equation}
\begin{equation}
F1_{avg} = \frac{F1_{Left} + F1_{Right}}{2}
\end{equation}

\subsubsection{Area Under ROC}
Finally, we used the area under ROC (AUROC) curve as our final evaluation metric. This metric evaluates the model performance in terms of classification separability.

\subsection{Classification Result}
Table \ref{tab:experiment_result} presents the performance of BERT-based model across both the target-independent and single-target stance detection tasks.

The performance of the BERT-based baseline model on the entire LASTIST dataset was not outstanding but remained understandable. 
Undoubtedly, the outcome shows the inherent complexity and contextual difficulty of the target-independent stance detection task, particularly when applied to a large-scale dataset such as the LASTIST. 

The performance of the single-target stance detection showed a significant improvement, outperforming other baselines. 
This outcome underscores the relative simplicity of the single-target detection task compared with the target-independent scenario. 
The observed performance gap further demonstrates the necessity for more advanced modeling strategies capable of capturing implicit stance in the absence of predefined target references and target-independent stance detection. 
Moreover, the strong performance of our BERT-based model reflects its ability to generate stance representations effectively, capturing affective contextual features from carefully annotated labels. 
This, in turn, confirms the quality and reliability of the dataset labels, implying their value as benchmarks for both single-target and target-independent stance detection tasks. 

\begin{table}[ht]
\centering
\begin{tabular}{l c c c}
\toprule
\textbf{Experiment Setting} & \textbf{Accuracy} & \textbf{F1} & \textbf{AUC-ROC} \\
\midrule
Target-Independent & 0.666 & 0.372 & 0.623 \\
Single-Target & 0.976 & 0.956 & 0.995 \\
\bottomrule
\end{tabular}
\caption{Performance comparison of BERT-based baseline under different experimental settings}
\label{tab:experiment_result}
\end{table}

\section{Discussion}
\subsection{Contribution of LASTIST dataset}
As shown in the experimental results, our study proposes a dataset construction framework and the validity of the new large-scale dataset that enables the training of target-independent stance classifiers, particularly in low-resource languages such as Korean.
Our classification results reveal that although frequently used SOTA models, such as BERT, excel at stance detection when the scope of the classification is restricted.
However, given that the same model cannot classify the stance when the scope of the dataset is expanded, it is reasonable to define a new task that enables stance classification even without an explicitly defined target.
In this context, our study proposes the LASTIST dataset, which can be used to train and evaluate a stance detection model designed to detect stances or biases that are not explicitly given. 

Moreover, our study introduces a new benchmark dataset that can facilitate model training in low-resource languages, such as Korean.
Although stance detection is a rapidly growing domain in ML for politics, most datasets focus on a couple of languages, such as English or Chinese, leading to a lack of multilingual support in stance detection research.
This lack of linguistic diversity limits the generalizability of existing stance detection models to other political contexts and languages.
Our study enables the development of a large-scale stance detection dataset from Korean political press releases to address this gap.
Therefore, our dataset is a valuable resource for evaluating cross-linguistic transferability and bias in stance classification tasks.

\subsection{Limitation}
Although this study introduced a new stance detection dataset for a large-scale target-independent task, it has several limitations.
First, using a subjectivity filter may not entirely reflect the nature of the stance represented in real life.
While subjectivity could serve as a valid filter to exclude sentences that take a stance toward an arbitrary target, it could also filter out sentences with a more implicit stance.
Modern studies on stance detection~\cite{hamborg2023revealing, 10266224} imply that while subjectivity could indicate the existence of a stance, it fails to provide a complete viewpoint on whether a stance exists in the data.
In other words, our dataset is limited because it does not include data that conveys objective information but has a specific stance.

Next, applying the active learning framework to auto-label the dataset might make its justification imperfect.
The use of an active learning framework was to respond to the amount of data collected, but its use still has limitations in terms of accuracy and reliability.
Our dataset is not fully constructed from manual labels; therefore, the labels might have internal biases that make them incomplete.
Further research with human annotations could follow our study by providing a more reliable and accurate labeling process for the collected data.

\section{Conclusion}
In this study, we introduce LASTIST, a Korean stance detection dataset designed to support large-scale target-independent stance detection.
Collected the press release articles from The Minjoo Party and People Power Party, our dataset consists of 
Our experimental results showed that target-independent stance detection tasks are much more complicated than target-specific detection, highlighting the need for further research on stance detection and machine learning applications in the political domain.

\appendix

\section{Data and Code availability}

We share our data and code for data collection, preprocessing, and experiment in our GitHub repository (\url{https://anonymous.4open.science/r/LASTIST-3721/}).

\bibliography{aaai2026}\

\begin{thebibliography}{39}
\providecommand{\natexlab}[1]{#1}

\bibitem[{Ajjour et~al.(2019)Ajjour, Wachsmuth, Kiesel, Potthast, Hagen, and
  Stein}]{10.1007/978-3-030-30179-8_4}
Ajjour, Y.; Wachsmuth, H.; Kiesel, J.; Potthast, M.; Hagen, M.; and Stein, B.
  2019.
\newblock Data Acquisition for Argument Search: The args.me Corpus.
\newblock In Benzm{\"u}ller, C.; and Stuckenschmidt, H., eds., \emph{KI 2019:
  Advances in Artificial Intelligence}, 48--59. Cham: Springer International
  Publishing.
\newblock ISBN 978-3-030-30179-8.

\bibitem[{ALDayel and Magdy(2021)}]{ALDAYEL2021102597}
ALDayel, A.; and Magdy, W. 2021.
\newblock Stance detection on social media: State of the art and trends.
\newblock \emph{Information Processing \& Management}, 58(4): 102597.

\bibitem[{Alturayeif, Luqman, and Ahmed(2023)}]{alturayeif2023systematic}
Alturayeif, N.; Luqman, H.; and Ahmed, M. 2023.
\newblock A systematic review of machine learning techniques for stance
  detection and its applications.
\newblock \emph{Neural Computing and Applications}, 35(7): 5113--5144.

\bibitem[{Antici et~al.(2024)Antici, Ruggeri, Galassi, Korre, Muti, Bardi,
  Fedotova, and Barr{\'o}n-Cede{\~n}o}]{antici-etal-2024-corpus}
Antici, F.; Ruggeri, F.; Galassi, A.; Korre, K.; Muti, A.; Bardi, A.; Fedotova,
  A.; and Barr{\'o}n-Cede{\~n}o, A. 2024.
\newblock A Corpus for Sentence-Level Subjectivity Detection on {E}nglish News
  Articles.
\newblock In Calzolari, N.; Kan, M.-Y.; Hoste, V.; Lenci, A.; Sakti, S.; and
  Xue, N., eds., \emph{Proceedings of the 2024 Joint International Conference
  on Computational Linguistics, Language Resources and Evaluation (LREC-COLING
  2024)}, 273--285. Torino, Italia: ELRA and ICCL.

\bibitem[{Bar-Haim et~al.(2017)Bar-Haim, Bhattacharya, Dinuzzo, Saha, and
  Slonim}]{bar-2017}
Bar-Haim, R.; Bhattacharya, I.; Dinuzzo, F.; Saha, A.; and Slonim, N. 2017.
\newblock Stance Classification of Context-Dependent Claims.
\newblock In Lapata, M.; Blunsom, P.; and Koller, A., eds., \emph{Proceedings
  of the 15th Conference of the {E}uropean Chapter of the Association for
  Computational Linguistics: Volume 1, Long Papers}, 251--261. Valencia, Spain:
  Association for Computational Linguistics.

\bibitem[{Caceres-Wright et~al.(2024)Caceres-Wright, Udhayasankar, Bunn,
  Shuster, and Joseph}]{10.1007/978-3-031-72241-7_1}
Caceres-Wright, A.~R.; Udhayasankar, N.; Bunn, G.; Shuster, S.~M.; and Joseph,
  K. 2024.
\newblock Explicit Stance Detection in the Political Domain: A New Concept
  and Associated Dataset.
\newblock In Thomson, R.; Hariharan, A.; Renshaw, S.; Al-khateeb, S.; Burger,
  A.; Park, P.; and Pyke, A., eds., \emph{Social, Cultural, and Behavioral
  Modeling}, 3--14. Cham: Springer Nature Switzerland.

\bibitem[{Choi, Shang, and Wang(2025)}]{10.1007/978-3-031-78538-2_28}
Choi, Y.; Shang, L.; and Wang, D. 2025.
\newblock ClimateMiSt: Climate Change Misinformation and Stance Detection
  Dataset.
\newblock In Aiello, L.~M.; Chakraborty, T.; and Gaito, S., eds., \emph{Social
  Networks Analysis and Mining}, 321--330. Cham: Springer Nature Switzerland.

\bibitem[{Conforti et~al.(2020)Conforti, Berndt, Pilehvar, Giannitsarou,
  Toxvaerd, and Collier}]{conforti-2020}
Conforti, C.; Berndt, J.; Pilehvar, M.~T.; Giannitsarou, C.; Toxvaerd, F.; and
  Collier, N. 2020.
\newblock Will-They-Won't-They: A Very Large Dataset for Stance Detection on
  {T}witter.
\newblock In Jurafsky, D.; Chai, J.; Schluter, N.; and Tetreault, J., eds.,
  \emph{Proceedings of the 58th Annual Meeting of the Association for
  Computational Linguistics}, 1715--1724. Online: Association for Computational
  Linguistics.

\bibitem[{Darwish, Magdy, and Zanouda(2017)}]{darwish-2017}
Darwish, K.; Magdy, W.; and Zanouda, T. 2017.
\newblock Trump vs. Hillary: What Went Viral During the 2016 US Presidential
  Election.
\newblock In \emph{Social Informatics}.

\bibitem[{Devlin et~al.(2019)Devlin, Chang, Lee, and
  Toutanova}]{devlin2019bert}
Devlin, J.; Chang, M.-W.; Lee, K.; and Toutanova, K. 2019.
\newblock Bert: Pre-training of deep bidirectional transformers for language
  understanding.
\newblock In \emph{Proceedings of the 2019 conference of the North American
  chapter of the association for computational linguistics: human language
  technologies, volume 1 (long and short papers)}, 4171--4186.

\bibitem[{Du~Bois(2008)}]{du2008stance}
Du~Bois, J.~W. 2008.
\newblock The stance triangle.
\newblock In \emph{Stancetaking in discourse: Subjectivity, evaluation,
  interaction}, 139--182. John Benjamins Publishing Company.

\bibitem[{Gera and Neal(2025)}]{gera2025deep}
Gera, P.; and Neal, T. 2025.
\newblock Deep Learning in Stance Detection: A Survey.
\newblock \emph{ACM Computing Surveys}.

\bibitem[{Gorrell et~al.(2019)Gorrell, Kochkina, Liakata, Aker, Zubiaga,
  Bontcheva, and Derczynski}]{gorrell-2019}
Gorrell, G.; Kochkina, E.; Liakata, M.; Aker, A.; Zubiaga, A.; Bontcheva, K.;
  and Derczynski, L. 2019.
\newblock {S}em{E}val-2019 Task 7: {R}umour{E}val, Determining Rumour Veracity
  and Support for Rumours.
\newblock In May, J.; Shutova, E.; Herbelot, A.; Zhu, X.; Apidianaki, M.; and
  Mohammad, S.~M., eds., \emph{Proceedings of the 13th International Workshop
  on Semantic Evaluation}, 845--854. Minneapolis, Minnesota, USA: Association
  for Computational Linguistics.

\bibitem[{Grimminger and Klinger(2021)}]{grimminger-2021}
Grimminger, L.; and Klinger, R. 2021.
\newblock Hate Towards the Political Opponent: A {T}witter Corpus Study of the
  2020 {US} Elections on the Basis of Offensive Speech and Stance Detection.
\newblock In De~Clercq, O.; Balahur, A.; Sedoc, J.; Barriere, V.; Tafreshi, S.;
  Buechel, S.; and Hoste, V., eds., \emph{Proceedings of the Eleventh Workshop
  on Computational Approaches to Subjectivity, Sentiment and Social Media
  Analysis}, 171--180. Online: Association for Computational Linguistics.

\bibitem[{Gyawali et~al.(2024)Gyawali, Sirbu, Sosea, Khanal, Caragea, Rebedea,
  and Caragea}]{gyawali-2024}
Gyawali, N.; Sirbu, I.; Sosea, T.; Khanal, S.; Caragea, D.; Rebedea, T.; and
  Caragea, C. 2024.
\newblock {G}un{S}tance: Stance Detection for Gun Control and Gun Regulation.
\newblock In Ku, L.-W.; Martins, A.; and Srikumar, V., eds., \emph{Proceedings
  of the 62nd Annual Meeting of the Association for Computational Linguistics
  (Volume 1: Long Papers)}, 12027--12044. Bangkok, Thailand: Association for
  Computational Linguistics.

\bibitem[{Hamborg(2023)}]{hamborg2023revealing}
Hamborg, F. 2023.
\newblock \emph{Revealing media bias in news articles: NLP techniques for
  automated frame analysis}.
\newblock Springer Nature.

\bibitem[{He et~al.(2024)He, Rao, Guo, Mokhberian, and
  Lerman}]{he-etal-2024-reading}
He, Z.; Rao, A.; Guo, S.; Mokhberian, N.; and Lerman, K. 2024.
\newblock Reading Between the Tweets: Deciphering Ideological Stances of
  Interconnected Mixed-Ideology Communities.
\newblock In Graham, Y.; and Purver, M., eds., \emph{Findings of the
  Association for Computational Linguistics: EACL 2024}, 1523--1536. St.
  Julian{'}s, Malta: Association for Computational Linguistics.

\bibitem[{Jim et~al.(2024)Jim, Talukder, Malakar, Kabir, Nur, and
  Mridha}]{JIM2024100059}
Jim, J.~R.; Talukder, M. A.~R.; Malakar, P.; Kabir, M.~M.; Nur, K.; and Mridha,
  M. 2024.
\newblock Recent advancements and challenges of NLP-based sentiment analysis: A
  state-of-the-art review.
\newblock \emph{Natural Language Processing Journal}, 6: 100059.

\bibitem[{Khouja(2020)}]{khouja-2020}
Khouja, J. 2020.
\newblock Stance Prediction and Claim Verification: An {A}rabic Perspective.
\newblock In Christodoulopoulos, C.; Thorne, J.; Vlachos, A.; Cocarascu, O.;
  and Mittal, A., eds., \emph{Proceedings of the Third Workshop on Fact
  Extraction and VERification (FEVER)}, 8--17. Online: Association for
  Computational Linguistics.

\bibitem[{Kim, Kim, and Park(2025)}]{kim2025know}
Kim, J.; Kim, D.; and Park, E. 2025.
\newblock I know your stance! Analyzing Twitter users’ political stance on
  diverse perspectives.
\newblock \emph{Journal of Big Data}, 12(1): 14.

\bibitem[{Kockelman(2004)}]{kockelman2004stance}
Kockelman, P. 2004.
\newblock Stance and subjectivity.
\newblock \emph{Journal of Linguistic Anthropology}, 14(2): 127--150.

\bibitem[{Kovacs, Cotfas, and Delcea(2024)}]{10.1007/978-3-031-70819-0_1}
Kovacs, E.-R.; Cotfas, L.-A.; and Delcea, C. 2024.
\newblock A Deep Learning Approach to Fine-Grained Political Ideology
  Classification on Social Media Texts.
\newblock In Nguyen, N.~T.; Franczyk, B.; Ludwig, A.; N{\'u}{\~{n}}ez, M.;
  Treur, J.; Vossen, G.; and Kozierkiewicz, A., eds., \emph{Computational
  Collective Intelligence}, 3--14. Cham: Springer Nature Switzerland.

\bibitem[{K{\"u}{\c{c}}{\"u}k and Can(2020)}]{kuccuk2020stance}
K{\"u}{\c{c}}{\"u}k, D.; and Can, F. 2020.
\newblock Stance detection: A survey.
\newblock \emph{ACM Computing Surveys (CSUR)}, 53(1): 1--37.

\bibitem[{Li et~al.(2021)Li, Sosea, Sawant, Nair, Inkpen, and
  Caragea}]{li2021p}
Li, Y.; Sosea, T.; Sawant, A.; Nair, A.~J.; Inkpen, D.; and Caragea, C. 2021.
\newblock P-stance: A large dataset for stance detection in political domain.
\newblock In \emph{Findings of the association for computational linguistics:
  ACL-IJCNLP 2021}, 2355--2365.

\bibitem[{Li and Zhang(2024)}]{li-2024}
Li, Y.; and Zhang, Y. 2024.
\newblock Pro-Woman, Anti-Man? Identifying Gender Bias in Stance Detection.
\newblock In Ku, L.-W.; Martins, A.; and Srikumar, V., eds., \emph{Findings of
  the Association for Computational Linguistics: ACL 2024}, 3229--3236.
  Bangkok, Thailand: Association for Computational Linguistics.

\bibitem[{Mets et~al.(2024)Mets, Karjus, Ibrus, and Schich}]{mets2024automated}
Mets, M.; Karjus, A.; Ibrus, I.; and Schich, M. 2024.
\newblock Automated stance detection in complex topics and small languages: the
  challenging case of immigration in polarizing news media.
\newblock \emph{Plos one}, 19(4): e0302380.

\bibitem[{Mohammad et~al.(2016)Mohammad, Kiritchenko, Sobhani, Zhu, and
  Cherry}]{mohammad2016semeval}
Mohammad, S.; Kiritchenko, S.; Sobhani, P.; Zhu, X.; and Cherry, C. 2016.
\newblock Semeval-2016 task 6: Detecting stance in tweets.
\newblock In \emph{Proceedings of the 10th international workshop on semantic
  evaluation (SemEval-2016)}, 31--41.

\bibitem[{Mohammad, Sobhani, and Kiritchenko(2017)}]{10.1145/3003433}
Mohammad, S.~M.; Sobhani, P.; and Kiritchenko, S. 2017.
\newblock Stance and Sentiment in Tweets.
\newblock \emph{ACM Trans. Internet Technol.}, 17(3).

\bibitem[{Niu et~al.(2024)Niu, Yang, Li, Zhang, Peng, and
  Zhang}]{niu2024challenge}
Niu, F.; Yang, M.; Li, A.; Zhang, B.; Peng, X.; and Zhang, B. 2024.
\newblock A challenge dataset and effective models for conversational stance
  detection.
\newblock \emph{arXiv preprint arXiv:2403.11145}.

\bibitem[{Rogers, Gardner, and Augenstein(2023)}]{10.1145/3560260}
Rogers, A.; Gardner, M.; and Augenstein, I. 2023.
\newblock QA Dataset Explosion: A Taxonomy of NLP Resources for Question
  Answering and Reading Comprehension.
\newblock \emph{ACM Comput. Surv.}, 55(10).

\bibitem[{Schr{\"o}der et~al.(2023)Schr{\"o}der, M{\"u}ller, Niekler, and
  Potthast}]{schroder-etal-2023-small}
Schr{\"o}der, C.; M{\"u}ller, L.; Niekler, A.; and Potthast, M. 2023.
\newblock Small-Text: Active Learning for Text Classification in Python.
\newblock In Croce, D.; and Soldaini, L., eds., \emph{Proceedings of the 17th
  Conference of the European Chapter of the Association for Computational
  Linguistics: System Demonstrations}, 84--95. Dubrovnik, Croatia: Association
  for Computational Linguistics.

\bibitem[{Sobhani, Inkpen, and Zhu(2017)}]{sobhani-etal-2017-dataset}
Sobhani, P.; Inkpen, D.; and Zhu, X. 2017.
\newblock A Dataset for Multi-Target Stance Detection.
\newblock In Lapata, M.; Blunsom, P.; and Koller, A., eds., \emph{Proceedings
  of the 15th Conference of the {E}uropean Chapter of the Association for
  Computational Linguistics: Volume 2, Short Papers}, 551--557. Valencia,
  Spain: Association for Computational Linguistics.

\bibitem[{Upadhyaya, Fisichella, and Nejdl(2023)}]{upadhyaya2023multi}
Upadhyaya, A.; Fisichella, M.; and Nejdl, W. 2023.
\newblock A multi-task model for emotion and offensive aided stance detection
  of climate change tweets.
\newblock In \emph{Proceedings of the ACM Web Conference 2023}, 3948--3958.

\bibitem[{Wang et~al.(2024)Wang, Zuo, Peng, and Plank}]{wang2024multiclimate}
Wang, J.; Zuo, L.; Peng, S.; and Plank, B. 2024.
\newblock MultiClimate: Multimodal Stance Detection on Climate Change Videos.
\newblock In \emph{Third Workshop on NLP for Positive Impact}, 315.

\bibitem[{Zhao and Caragea(2024)}]{ezstance-2024}
Zhao, C.; and Caragea, C. 2024.
\newblock {EZ}-{STANCE}: A Large Dataset for {E}nglish Zero-Shot Stance
  Detection.
\newblock In Ku, L.-W.; Martins, A.; and Srikumar, V., eds., \emph{Proceedings
  of the 62nd Annual Meeting of the Association for Computational Linguistics
  (Volume 1: Long Papers)}, 15697--15714. Bangkok, Thailand: Association for
  Computational Linguistics.

\bibitem[{Zhao, Li, and Caragea(2023)}]{cstance-2023}
Zhao, C.; Li, Y.; and Caragea, C. 2023.
\newblock {C}-{STANCE}: A Large Dataset for {C}hinese Zero-Shot Stance
  Detection.
\newblock In Rogers, A.; Boyd-Graber, J.; and Okazaki, N., eds.,
  \emph{Proceedings of the 61st Annual Meeting of the Association for
  Computational Linguistics (Volume 1: Long Papers)}, 13369--13385. Toronto,
  Canada: Association for Computational Linguistics.

\bibitem[{Zhao, Wang, and Peng(2023)}]{orchid-2023}
Zhao, X.; Wang, K.; and Peng, W. 2023.
\newblock {ORCHID}: A {C}hinese Debate Corpus for Target-Independent Stance
  Detection and Argumentative Dialogue Summarization.
\newblock In Bouamor, H.; Pino, J.; and Bali, K., eds., \emph{Proceedings of
  the 2023 Conference on Empirical Methods in Natural Language Processing},
  9358--9375. Singapore: Association for Computational Linguistics.

\bibitem[{Zhukova et~al.(2023)Zhukova, Ruas, Hamborg, Donnay, and
  Gipp}]{10266224}
Zhukova, A.; Ruas, T.; Hamborg, F.; Donnay, K.; and Gipp, B. 2023.
\newblock What's in the News? Towards Identification of Bias by Commission,
  Omission, and Source Selection (COSS).
\newblock In \emph{2023 ACM/IEEE Joint Conference on Digital Libraries (JCDL)},
  258--259.

\bibitem[{Zotova et~al.(2020)Zotova, Agerri, Nu{\~n}ez, and
  Rigau}]{zotova-2020}
Zotova, E.; Agerri, R.; Nu{\~n}ez, M.; and Rigau, G. 2020.
\newblock Multilingual Stance Detection in Tweets: The {C}atalonia Independence
  Corpus.
\newblock In Calzolari, N.; B{\'e}chet, F.; Blache, P.; Choukri, K.; Cieri, C.;
  Declerck, T.; Goggi, S.; Isahara, H.; Maegaard, B.; Mariani, J.; Mazo, H.;
  Moreno, A.; Odijk, J.; and Piperidis, S., eds., \emph{Proceedings of the
  Twelfth Language Resources and Evaluation Conference}, 1368--1375. Marseille,
  France: European Language Resources Association.
\newblock ISBN 979-10-95546-34-4.

\end{thebibliography}

\end{document}